\UseRawInputEncoding
\documentclass[letterpaper]{article} 
\usepackage{aaai23}  
\usepackage{times}  
\usepackage{helvet}  
\usepackage{courier}  
\usepackage[hyphens]{url}  
\usepackage{graphicx} 
\usepackage{natbib}  
\usepackage{caption} 
\usepackage{algorithm}
\usepackage[noend]{algorithmic}

\usepackage{newfloat}
\usepackage{listings}

\usepackage{soul}
\usepackage{subfigure}
\usepackage{multirow}
\usepackage[normalem]{ulem}
\usepackage{booktabs}
\usepackage{amsthm,amsmath,amssymb}
\usepackage{mathrsfs}
\usepackage{arydshln}
\usepackage{xr}
\usepackage{eqparbox}
\usepackage{threeparttable}

\useunder{\uline}{\ul}{}

\newtheorem{myDef}{Definition}
\usepackage[inline]{enumitem}   
\makeatletter
\newcommand{\inlineitem}[1][]{%
\ifnum\enit@type=\tw@
    {\descriptionlabel{#1}}
  \hspace{\labelsep}%
\else
  \ifnum\enit@type=\z@
       \refstepcounter{\@listctr}\fi
    \quad\@itemlabel\hspace{\labelsep}%
\fi}
\makeatother

\makeatletter
\newcommand*{\addFileDependency}[1]{
  \typeout{(#1)}
  \@addtofilelist{#1}
  \IfFileExists{#1}{}{\typeout{No file #1.}}
}
\makeatother

\newcommand*{\myexternaldocument}[1]{%
    \externaldocument{#1}%
    \addFileDependency{#1.tex}%
    \addFileDependency{#1.aux}%
}

\myexternaldocument{supplementary-material}

\DeclareCaptionStyle{ruled}{labelfont=normalfont,labelsep=colon,strut=off} 
\floatstyle{ruled}
\newfloat{listing}{tb}{lst}{}
\floatname{listing}{Listing}
\title{Multi-Aspect Explainable Inductive Relation Prediction by Sentence Transformer}

\author{
    Zhixiang Su \textsuperscript{\rm 1,2},
    Di Wang \textsuperscript{\rm 3,4}, 
    Chunyan Miao \textsuperscript{\rm 1,2,3,4}, 
    Lizhen Cui\textsuperscript{\rm 2,5}
}
\affiliations{
    \textsuperscript{\rm 1}School of Computer Science and Engineering, Nanyang Technological University (NTU), Singapore\\
    \textsuperscript{\rm 2}SDU-NTU Centre for Artificial Intelligence Research (C-FAIR), Shandong University (SDU), China\\
    \textsuperscript{\rm 3}Joint NTU-UBC Research Centre of Excellence in Active Living for the Elderly (LILY), NTU, Singapore\\
    \textsuperscript{\rm 4}Joint NTU-WeBank Research Centre on Fintech, NTU, Singapore\\
    \textsuperscript{\rm 5}School of Software, SDU, China\\ 
    \{zhixiang002, wangdi, ascymiao\}@ntu.edu.sg, clz@sdu.edu.cn 

}

\usepackage{bibentry}
\nocopyright
\begin{document}
\maketitle

\begin{abstract}
    Recent studies on knowledge graphs (KGs) show that path-based methods empowered by pre-trained language models perform well in the provision of inductive and explainable relation predictions. In this paper, we introduce the concepts of relation path coverage and relation path confidence to filter out unreliable paths prior to model training to elevate the model performance. Moreover, we propose Knowledge Reasoning Sentence Transformer (KRST) to predict inductive relations in KGs. KRST is designed to encode the extracted reliable paths in KGs, allowing us to properly cluster paths and provide multi-aspect explanations. We conduct extensive experiments on three real-world datasets. The experimental results show that compared to SOTA models, KRST achieves the best performance in most transductive and inductive test cases (4 of 6), and in 11 of 12 few-shot test cases.
\end{abstract}

\section{Introduction}\label{section_introduction}
    As an important tool for providing side information for question answering and recommendation systems~\cite{ji2021survey}, knowledge graph (KG) has been widely studied. A KG is typically expressed in terms of triplets $G=\{(h_i,r_i,t_i)|i=1,2,3,..,m\}$, which contains entities $h_i,t_i \in E_G$ and relations $r_i \in R_G$. Because of the incompleteness of KGs in practice, knowledge graph completion (KGC) is needed to improve the quality of KGs. One of the most important KGC tasks is relation prediction. Given the target triplet $(h,r,t)$, a relation prediction query is usually set by masking the entity $h$ or $t$ in the given triplet and letting the model predict the masked entity based on the other entity and the relation type.
    
    Embedding-based methods are probably the most commonly applied SOTA models.  With a fixed set of entities and relations, embedding-based methods perform fairly well in KGC tasks. However, most existing embedding-based methods are not explainable and cannot deal with inductive situations, making them not suitable for modeling real-world dynamic KGs, wherein new entities and relations may be added all the time. Inductive relation prediction requires the model to handle unseen entities unavailable in the training graph. 
    
    GNN-based methods take advantage of the KG's graph connectivity, thus, are capable of predicting new entities with a sufficient number of known neighboring entities. Recent GNN-based method GRAIL~\cite{GRAIL} is shown to be capable of conducting inductive relation predictions. Nonetheless, extracting explainable rules is left for further exploration in GRAIL. So far, no evidence shows that GRAIL is explainable.

    \begin{figure}[!t]
        \centering
        \includegraphics[scale=0.62]{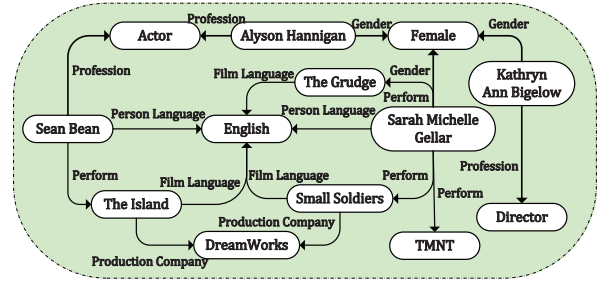}
        \caption{An example of KG.}\label{fig2.0}
    \end{figure}

    Different from GNN-based methods, path-based methods take advantage of the graph connectivity by analyzing paths between the head and tail entities. When we focus only on the relations in one path, they can be formulated as a Horn rule \cite{HornRule} for knowledge reasoning. Therefore, in path-based methods, new entities can be easily modeled by applying the summarized rules. 
    For example, w.r.t Figure~\ref{fig2.0}, for triplet $(\textit{Small Soldiers}, \textit{Film Language}, \textit{English})$ and the corresponding path 
    \begin{equation}
    \scriptsize
        \textit{Small Soldiers} \xrightarrow[]{\textit{Perform}^{\textit{-1}}} \textit{Sarah} \xrightarrow[]{\textit{Person Language}} \textit{English},
    \end{equation}
    we can summarize that
    \begin{equation}
    \scriptsize
        \begin{aligned}
            (y, \textit{Perform}, x) \wedge (y, \textit{PersonLanguage}, z) 
            \rightarrow (x, \textit{FilmLanguage}, z).
        \end{aligned}
    \end{equation}
    Such rule-based operations make path-based methods inductive and highly explainable. BERTRL~\cite{BERTRL}, which employs the pre-trained language model BERT for scoring, is one typical model of such type. Using contextual descriptions of entities and relations, BERTRL can deal with inductive cases and provide single-path explanations, achieving the best SOTA results in the literature.

    To further improve the relation prediction performance and strive for better explainability, in this paper, we propose \textbf{Knowledge Reasoning Sentence Transformer (KRST)}, which is a novel path-based model built upon the sentence transformer. The key innovations of KRST are as follows:
    
    \noindent \textbf{Path extraction.} Although paths between head and tail entities can be easily extracted, most of the extracted paths may be unreliable. Moreover, unreliable short paths with words also appearing in the target triplets are usually preferred by the pre-trained language model. To assess the reliability of paths and only use reliable ones for model training, we propose the concepts of relation path coverage and relation path confidence (see Definitions~\ref{def_relation_path_coverage} and~\ref{def_relation_path_confidence}, respectively). 

    \noindent \textbf{Pre-trained language model.} Compared with the commonly applied BERT for sequence classification model, the sentence transformer adopting cosine similarity achieves a higher performance in multiple tasks  \cite{SentenceTransformer}. Therefore, we adopt sentence transformer in KRST to encode triplets and paths into embeddings, which allows us to explicitly compare between embeddings and cluster paths w.r.t various aspects. 
        
    \noindent \textbf{Loss function.} In KRST, we compare the similarity between paths and triplets using the cosine similarity score. In the case of negative triplets, negative scores are not necessarily close to $-1$. Nevertheless, commonly applied binary classification loss functions (e.g., cross-entropy) often make the model over-confident by requiring the prediction result to be close to either $1$ or $-1$. Therefore, we use cosine embedding loss instead, aiming to better elevate the model performance.
    
    Our key contributions in this paper are as follows:
    
    (\romannumeral1) We propose two novel path extraction metrics named relation path coverage and relation path confidence, and a novel path-based model named KRST. To the best of our knowledge, KRST is the first sentence transformer model for knowledge graph path encoding.

    (\romannumeral2) We develop a comprehensive approach for relation prediction explanation, which enables the provision of explanations from multiple paths and multiple perspectives.
    
    (\romannumeral3) We assess the performance of KRST on three transductive and inductive datasets: WN18RR, FB15k-237, and NELL-995. KRST obtains significantly better results than SOTA models in majority cases (15 of 18).


\section{Related Work}\label{section_related_work}
    \textbf{Embedding-based methods:} Such methods (e.g., ComplEx~\cite{ComplEx}, ConvE~\cite{ConvE}, and TuckER~\cite{Tucker}) generate embeddings for entities and relations in the latent space. Score functions are proposed for training and evaluating within triplets. The most representative embedding-based methods are the translation methods (e.g., TransE~\cite{TransE}, TransH~\cite{TransH}, TransR~\cite{TransR}, and TransD~\cite{TransD}). The key idea behind the translation models is to treat the process of finding valid triplets as the translation operation of entities through relationships, define the corresponding score function, and then minimize the loss function to learn the representation of entities and relationships ~\cite{chen2020knowledge}.
    
    \noindent \textbf{GNN-based methods:} Such methods (e.g., CompGCN~\cite{CompGCN} and R-GCN~\cite{RGCN}) pass messages between a node and its neighbors. These approaches take advantage of the graph connectivity. They are able to deal with a particular inductive situation where the new entity is surrounded by entities already known. GRAIL~\cite{GRAIL} is proposed to handle KGs with entirely new entities. However, GRAIL is not explainable as the author stated in \cite{GRAIL}. Challenged by the number of reachable entities that grows exponentially with the search depth, when given a dense KG, GNN-based methods may not well capture the correct long path information and hence may not perform well in relation prediction tasks.
    
    \noindent \textbf{Path-based methods:} Such methods aim to find one (or multiple) logical reasoning path(s) between the query head and tail entities. PRA~\cite{PRA} and AMIE~\cite{AMIE} generate Horn rules~\cite{HornRule}, which have a broader definition than paths. However, due to noises in real-world KGs, their performance is limited because they are only applicable for exact matches. DeepPath~\cite{DeepPath} and MINERVA~\cite{MINERVA} learn to generate paths by reinforcement learning, whereby positive rewards are given when having successful target arrivals. These approaches can be applied to new entities and are naturally explainable. Nevertheless, they are also challenged by the exponentially growing reachable entities, which leads to sparse rewards.
    
    \noindent \textbf{Methods with pre-trained language model:} With the great success achieved in various NLP tasks, pre-trained language models (PLMs) (e.g., BERT~\cite{BERT}, GPT~\cite{GPT} and XLNet~\cite{Xlnet}) show great potential in dealing with contextual descriptions. KG-BERT~\cite{KGBERT} extends embedding-based methods by fine-tuning BERT. Using contextual descriptions for entities and relations, BERT model is able to understand a triplet and output a classification label. KG-BERT works well in inductive settings but is not explainable because of the incomprehensible embeddings. Different from KG-BERT, BERTRL~\cite{BERTRL} incorporates path-based methods with BERT. Specifically, BERTRL converts triplets and the corresponding paths into sentences, and fine-tunes the pre-trained BERT for sequence classification. Because BERTRL uses all paths (shorter than $L$) as inputs without filtering, its performance may be limited. Recently, sentence transformer models is shown to outperform BERT on common STS and transfer learning tasks~\cite{SentenceTransformer}. In addition, BERT for sequence classification requires two sentences to input together and make an implicit comparison, while sentence transformer encodes sentences separately, allowing more flexible comparisons among sentences. In this paper, we adopt sentence transformer to provide a multi-aspect explanation.

\section{Preliminary}\label{section_preliminary}
    We start this section by introducing the commonly applied logical reasoning path. Then we introduce the definition and common settings for inductive relation prediction.
    
    \begin{myDef}[Logical Reasoning Path]
    \label{def_logical_reasoning_path}
    Given a KG $G=\{(h_i,r_i,t_i)|i=1,2,3,..,m\}$, $h_i,t_i \in E_G$ and $r_i \in R_G$, one possible logical reasoning path $p(h,r,t)$ and the corresponding relation path $R_p(h,r,t)$ are defined as follows:
    \begin{equation}
    \footnotesize
        p(h,r,t)= h \xrightarrow[]{r_1} e_1 \xrightarrow[]{r_2} e_2 \xrightarrow[]{r_3} ... \xrightarrow[]{r_{n-1}} e_{n-1} \xrightarrow[]{r_n} t,
    \end{equation}
    \begin{equation}
    \footnotesize
        R_p(h,r,t)=(r_1,r_2,...,r_{n-1},r_n),
    \end{equation}
    where $(h,r_1,e_1),...,(e_{n-1},r_n,t) \in G-\{(h,r,t)\}$.
    \end{myDef}

    From the perspective of knowledge graph reasoning, the relation prediction task can be viewed as a logical induction problem to identify the inductive and explainable logical reasoning paths. Because logical reasoning paths are sequential in nature, we can easily convert them into sentences.
    However, a large proportion of paths are either illogical or entity-dependent in real-world KGs, which cannot be applied to inductive relation prediction (see Definition~\ref{def_inductive_relation_prediction}). To address this problem, we define relation path coverage and relation path confidence to filter out unreliable or meaningless paths (see Definitions~\ref{def_relation_path_coverage} and~\ref{def_relation_path_confidence}, respectively).

    \begin{myDef}[Inductive Relation Prediction]
    \label{def_inductive_relation_prediction}
    Given a training graph $G_{\textit{train}}(E_{G_{\textit{train}}},R_{G_{\textit{train}}})$, a testing graph $G_{\textit{test}}(E_{G_{\textit{test}}},R_{G_{\textit{test}}})$ and a query triplet $(h_q,r_q,t_q)$,  a relation prediction is inductive if:
    \begin{itemize}
        \item $E_{G_{\textit{train}}} \cap E_{G_{\textit{test}}}= \emptyset$,
        \item $R_{G_{\textit{test}}} \subseteq R_{G_{\textit{train}}}, r_q \in R_{G_{\textit{train}}}$.
    \end{itemize}
    \end{myDef}
    With the emergence of real-world ever-evolving KGs, dealing with new relations and entities is a necessity. We focus on the inductive setting introduced by GRAIL~\cite{GRAIL}, which contains a training graph, a testing graph, and a series of query triplets. Only the training graph is visible during training. Because both GNN-based methods and path-based methods make use of the graph connectivity, the test graph is only applied to extract neighbors or paths w.r.t query triplets during testing, respectively.

\section{Methodology}\label{section_methodology}
    Our proposed architecture comprises three steps for relation prediction, with the intuition that a triplet and its corresponding reliable paths should have similar semantics when converted to sentences. Specifically, KRST 1) filters unreliable logical reasoning paths extracted for model training, 2) converts paths and triplets into sentences by sentence formation, and 3) measures semantic similarity scores and makes relation predictions based on them.
\begin{algorithm}[tbh!]
            \caption{Head Relation Path Coverage}
            \label{alg_head_relation_path_coverage}
            \textbf{Input}: KG $G$, query triplet $(h,r,t)$, relation path length $|R_{p^{'}}|$\\
            \textbf{Output}: Coverage rate $\textit{cover}$
            \begin{algorithmic}[1] 
                \STATE \COMMENT{Create a nested dictionary}
                \STATE \COMMENT{Dimension 1: entity name}
                \STATE \COMMENT{Dimension 2: depth}
                \STATE $\textit{pathNum}=\textit{Dictionary}(\textit{Dictionary}())$; \STATE $\textit{count}=0$; $\textit{support}=0$
                \STATE \COMMENT{Iterately update the dictionary}
                \FOR{$i \in \textit{range}(0,|R_{p^{'}}|)$}
                    \STATE $\textit{newNum}=\textit{pathNum}$.copy()
                    \FOR{$e \in \textit{pathNum}$}
                        \FOR{$v \in G[e]$}
                            \FOR{$\textit{dpth} \in \textit{pathNum}[e]$}
                                \STATE $\textit{newNum}[v][dpth+1]=\textit{pathNum}[e][dpth]$
                            \ENDFOR
                        \ENDFOR
                    \ENDFOR
                    \STATE $\textit{pathNum}=\textit{newNum}$.copy()
                \ENDFOR
                \STATE \COMMENT{Sum up number of paths}
                \FOR{$e \in \textit{pathNum}$}
                    \IF{$e==t$}
                        \STATE $\textit{support}+=\textit{pathNum}[e][|R_{p^{'}}|]$
                    \ENDIF
                    \STATE $\textit{count}+=\textit{pathNum}[e][|R_{p^{'}}|]$
                \ENDFOR
                \STATE $\textit{cover}=\textit{support}/\textit{count}$
                \STATE \textbf{return} $\textit{cover}$
            \end{algorithmic}
        \end{algorithm}
        
        \begin{algorithm}[tbh!]
            \caption{Head Relation Path Confidence}
            \label{alg_head_relation_path_confidence}
            \textbf{Input}: KG $G$, query triplet $(h,r,t)$, logical reasoning path $P$, relation path $R_{p^{'}}$\\
            \textbf{Output}: Confidence score $\textit{conf}$
            \begin{algorithmic}[1] 
                \STATE $q=\textit{Queue}()$
                \STATE $q$.push($(h,0)$); $\textit{count}=0$; $\textit{support}=0$
                \STATE \COMMENT{Breadth-first search} 
                \WHILE{$q$ is not empty}
                    \STATE $u,l=q$.pop()
                    \IF {$l==|R_{p^{'}}|$}
                        \IF {$u==t$}
                            \STATE $\textit{support}=\textit{support}+1$
                        \ENDIF
                        \STATE $\textit{count}=\textit{count}+1$
                        \STATE \textbf{continue}
                    \ENDIF
                    \FOR{$v$ in $G[u]$}
                        \STATE \COMMENT{Check relation path}
                        \IF{$G[u][v][\textit{'relation'}]==R_{p^{'}}[l]$}
                            \STATE $q$.push($(v,l+1)$)
                        \ENDIF
                    \ENDFOR
                \ENDWHILE
                \STATE $\textit{conf}=\textit{support}/\textit{count}$
                \STATE \textbf{return} $\textit{conf}$
            \end{algorithmic}
        \end{algorithm}

    \subsection{Path Extraction with Filtering}
        
    Following the idea of knowledge reasoning, a logical reasoning path is determined by its support evidence w.r.t the target triplet. To train the model with both positive and negative samples, we extract logical reasoning paths from KGs for both positive and negative target triplets. However, many extracted paths are unreliable. For example, for positive triplet $(\textit{Sarah}, \textit{Profession}, \textit{Actor})$ and negative triplet $(\textit{Sarah}, \textit{Profession}, \textit{Director})$, we can both extract paths with the same relation path $(\textit{Gender},\textit{Gender}^{\textit{-1}},\textit{Profession})$ as follows:
        \begin{equation}
        \label{path3}
        \scriptsize
            \textit{Sarah} \xrightarrow[]{\textit{Gender}} \textit{Female} \xrightarrow[]{\textit{Gender}^{\textit{-1}}} \textit{Kathryn} \xrightarrow[]{\textit{Profession}} \textit{Actor},
        \end{equation}
        \begin{equation}
        \label{path3.1}
        \scriptsize
            \textit{Sarah} \xrightarrow[]{\textit{Gender}} \textit{Female} \xrightarrow[]{\textit{Gender}^{\textit{-1}}} \textit{Alyson} \xrightarrow[]{\textit{Profession}} \textit{Director}.
        \end{equation}
    
    \noindent Paths~(\ref{path3}) and (\ref{path3.1}) are similar and contribute little to distinguish the profession. Therefore, we consider paths with relation path $(\textit{Gender},\textit{Gender}^{\textit{-1}},\textit{Profession})$ to be unreliable. Although the unreliable paths should be excluded from model training, because they are similar to the target triplet, they may be given a high similarity score and hence mistakenly considered as reliable. For example, another path for triplet $(\textit{Sarah}, \textit{Profession}, \textit{Actor})$ is as follows:
        \begin{equation}
        \label{path4}
        \scriptsize
            \begin{aligned}
                \textit{Sarah} \xrightarrow[]{\textit{Perform}} \textit{SmallSoldiers} \xrightarrow[]{\textit{Company}^{\textit{-1}}} 
                \textit{DreamWorks} \\ \xrightarrow[]{\textit{Company}}  \textit{TheIsland} \xrightarrow[]{\textit{Perform}^{\textit{-1}}} \textit{Sean} \xrightarrow[]{\textit{Profession}} \textit{Actor}.
            \end{aligned}
        \end{equation}
    This path is relatively more reliable than Path~(\ref{path3}) because it involves the person in the same company for prediction. However, after being converted to sentences, Path~(\ref{path3}) is assumed to be more similar by PLMs (shorter and contains words in target triplet). Because $(\textit{Sarah}, \textit{Profession}, \textit{Actor})$ is a positive triplet, model training reinforces this bias, leading to an even higher score for Path~(\ref{path3}) after using it for training. These unreliable paths significantly limit the model performance. To exclude unreliable extracted paths from training, we perform path filtering.
    How we identify unreliable logical reasoning paths are introduced as follows.
    
    \begin{myDef}[Relation Path Support]
        \label{def_relation_path_support}
        Given a triplet $(h,r,t)$ and a relation path $R_{p^{'}}=(r^{'}_{1},r^{'}_{2},...,r^{'}_{m-1},r^{'}_{m})$, the support of $R_{p^{'}}$ is defined as follows:
        \begin{equation}
        \footnotesize
            \textit{supp}_{R_{p^{'}}} (h,r,t) = \#p(h,r,t), R_p(h,r,t)=R_{p^{'}},
        \end{equation}
        where $\#p(h,r,t)$ denotes the number of logical reasoning paths on $G$ w.r.t $(h,r,t)$. 
    \end{myDef}
    Relation path support $\textit{supp}_{R_{p^{'}}} (h,r,t)$ measures the number of paths containing the same relations with $R_{p^{'}}$ between $h$ and $t$. With a larger support score, $R_{p^{'}}$ is more common among paths between $h$ and $t$. 
    However, relation path support represents an unbounded value. Entity pairs with better connectivity usually have a larger support number. To further assess the ratio of the support, we define relation path coverage and relation path confidence, respectively.

    \begin{myDef}[Relation Path Coverage]
        \label{def_relation_path_coverage}
        Given a triplet $(h,r,t)$ and a relation path $R_{p^{'}}$, the head and tail relation path coverage of $R_{p^{'}}$ is defined as follows:
        \begin{equation}\label{eq9}
        \footnotesize
            \textit{cover}_{R_{p^{'}}}^{h} (h,r,t) = \frac{\textit{supp}_{R_{p^{'}}} (h,r,t)}{\#p(h,r,t^{'}), |p(h,r,t^{'})|=|R_{p^{'}}|},
        \end{equation}
        \begin{equation}\label{eq10}
        \footnotesize
            \textit{cover}_{R_{p^{'}}}^{t} (h,r,t) = \frac{\textit{supp}_{R_{p^{'}}} (h,r,t)}{\#p(h^{'},r,t), |p(h^{'},r,t)|=|R_{p^{'}}|},
        \end{equation}
        where $|\cdot|$ denotes the length of the path.
    \end{myDef}
    In (\ref{eq9}) and (\ref{eq10}), the number of paths starting from $h$ or ending at $t$ with length $|R_{p^{'}}|$ is adopted as the denominator, respectively. 
    \begin{myDef}[Relation Path Confidence]
        \label{def_relation_path_confidence}
        Given a triplet $(h,r,t)$, the head and tail relation path confidence of a relation path $R_{p^{'}}$ is defined as follows:
        \begin{equation}\label{eq11}
        \footnotesize
            \textit{conf}_{R_{p^{'}}}^{h} (h,r,t) = \frac{\textit{supp}_{R_{p^{'}}} (h,r,t)}{\sum_{t' \in E_G-\{h\}} \textit{supp}_{R_{p^{'}}} (h,r,t')},
        \end{equation}
        \begin{equation}\label{eq12}
        \footnotesize
            \textit{conf}_{R_{p^{'}}}^{t} (h,r,t) = \frac{\textit{supp}_{R_{p^{'}}} (h,r,t)}{\sum_{h' \in E_G-\{t\}} \textit{supp}_{R_{p^{'}}} (h',r,t)}.
        \end{equation}
    \end{myDef}
    In (\ref{eq11}) and (\ref{eq12}), the total number of all reachable entities starting from $h$ or ending at $t$ w.r.t relation path $R_{p^{'}}$ is adopted as the denominator, respectively.


    \begin{algorithm}[tbh!]
            \caption{Path Extraction}
            \label{alg_path_extraction}
            \textbf{Input}: KG $G$, query triplet $(h,r,t)$, filter threshold $\alpha$, filter function $f()$, max search depth $L$, max number of paths $M$ \\
            \textbf{Output}: List of extracted paths $P$
            \begin{algorithmic}[1]
                \STATE \COMMENT{Initialize search queue and state list}
                \STATE $q=\textit{Queue}()$;  $\textit{visited}=\textit{List}()$; $\textit{prev}=\textit{List}()$ 
                \STATE $q$.push($(h,0)$)
                \STATE $\textit{visited}[h]=\textit{True}$
                \STATE \COMMENT{Breadth-first search}
                \WHILE{$q$ is not empty}
                    \STATE $u,l=q$.pop()
                    \STATE \COMMENT{Check whether search depth exceeds $L$}
                    \IF{$l>=L$}
                        \STATE \textbf{continue}
                    \ENDIF
                    \STATE \COMMENT{Give priority to less frequent relations}
                    \FOR{$v$ in $G[u]$ sorted by frequency $G[u][v][\textit{`relation'}]$}
                        \IF{$v==t$}
                            \IF{$u==h \&\& G[u][v][\textit{`relation'}]==r$}
                                \STATE \textbf{continue}
                            \ENDIF
                            \STATE $p=$generatePath($\textit{prev},h,t$)
                            \IF{$f(p) \geq \alpha$}
                                \STATE $P$.add($p$)
                            \ENDIF  
                        \ELSIF{$\textit{visited}[v]==0$}
                            \STATE $q$.push($(v,l+1)$); $\textit{visited}(v)=1$; $\textit{prev}(v)=u$
                        \ENDIF
                        \COMMENT{Early break when generated path is enough}
                        \IF{$|P|>M$}
                            \STATE \textbf{break}
                        \ENDIF
                    \ENDFOR
                    \IF{$|P|>M$}
                            \STATE \textbf{break}
                    \ENDIF
                \ENDWHILE
                \STATE \textbf{return} $P$
            \end{algorithmic}
        \end{algorithm}
    \begin{figure*}[tbh!]
            \centering
            \includegraphics[scale=0.5]{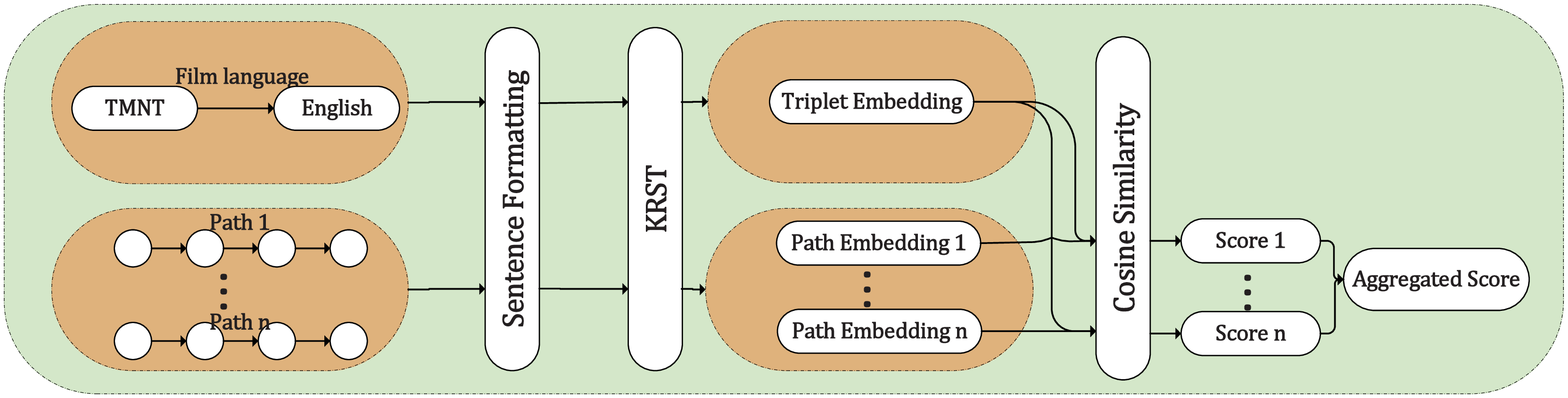}
            \caption{Overall architecture of relation prediction in KRST.}\label{fig5}
    \end{figure*} 
    
    Relation path coverage measures the ratio over all paths with the same source (or destination) and length, while relation path confidence measures the ratio over all entities that satisfy the target relation paths. To show the effectiveness of relation path confidence, we 
    refer back to Paths~(\ref{path3}) and~(\ref{path4}). The relation path confidence score for Path~(\ref{path4}) is much larger than Path~(\ref{path3}). Because in the relation path of $(\textit{Gender},\textit{Gender}^{\textit{-1}},\textit{Profession})$, $\textit{Gender}^{\textit{-1}}$ can lead to multiple people with various professions. Therefore, the number of paths satisfying the relation path of Path~(\ref{path3}) is significantly more than that of Path~(\ref{path4}), leading to a relatively smaller relation path confidence score for Path~(\ref{path3}).The implementations of these two metrics in head cases are presented in Algorithms~\ref{alg_head_relation_path_coverage} and \ref{alg_head_relation_path_confidence}, respectively. They can be easily converted to tail cases using reversed graphs.

    For path extraction, we adopt the breadth-first search, with the maximum search depth $L$ and the maximum number of paths per triplet $M$. These parameters are set to avoid the generation of an unnecessarily large number of paths. Also, paths with excessive length are highly likely illogical. In addition, to ensure a sufficient number of paths per triplet are generated for effective model training, we synchronously perform path filtering and path extraction (see Algorithm~\ref{alg_path_extraction}).

    \subsection{Sentence Formation}

    To leverage the pre-trained parameters of the sentence transformer, KRST converts paths and triplets into sentences. Our key considerations are as follows:
    
    \noindent \textbf{Entity description selection.} Both long (more than 20 words on average) and short entity descriptions in text are available in various datasets. Long descriptions usually make the sentence description imbalance between entities and relations when being modeled by PLMs, hence, they are less effective under inductive situations. In addition, sentences converted using long descriptions usually exceed the maximum sequence length of the PLM, which need to be truncated as incomplete. Therefore, only short descriptions are applied to KRST.
    
    \noindent \textbf{Inverse relation.} To provide KRST with sequential order for positional encoding, entities' and relations' order in the paths should be preserved after being converted into sentences. A straightforward way is to place the descriptions following the entities' and relations' order in paths. However, inverse relations (e.g., $\textit{Gender}^{\textit{-1}}$ in Path~(\ref{path3})) occur in the paths and their descriptions are not available. Empirically, a systematic description in KG usually starts with descriptions related to the head entity and ends with descriptions related to the tail entity. Thus, we choose to use the inverse order of words in the sequential relation for inverse relations. Moreover, by doing so, we preserve the similarity between relations and inverse ones, making it easier for the model to understand symmetric relations (e.g., $\textit{friend}$, $\textit{spouse}$, and $\textit{teammate}$).
    
    \noindent \textbf{Description concatenation.} To obtain a complete sentence, we need to concatenate descriptions of entities and relations in the path. A natural language pattern makes the sentence closer to human expressions. An example w.r.t triplet $(\textit{Sarah}, \textit{Profession}, \textit{Actor})$  is shown as follows:
    \begin{equation*}
    \scriptsize
        \begin{aligned}
        & \textit{Question:}  \\ 
        & \textit{\textbf{Sarah Michelle Gellar} is the \textbf{person profession} of what?} \\
        & \textit{Is the correct answer \textbf{Actor}?} 
        \end{aligned}
    \end{equation*}
    However, our preliminary results show that formulating complete sentences does not yield better performance than simply combining entities and relations together using semicolons as follows:
    \begin{equation*}
    \scriptsize
        \textit{Sarah Michelle Gellar; person profession; Actor}
    \end{equation*}
    This is because the latter approach better preserves the sequential order and relative positions in the paths in PLMs. Therefore, we choose to use semicolons for description concatenation.
    For example, the corresponding sentence of Path~(\ref{path3}) is formulated as follows:
    \begin{equation*}
    \scriptsize
        \begin{aligned}
        & \textit{Sarah Michelle Gellar; person gender; Female;}  \\ 
        & \textit{ gender person; Alyson Hannigan; person profession; Actor}
        \end{aligned}
    \end{equation*}

    \subsection{KRST Prediction}

    After sentence formation, KRST is able to generate embeddings and make relation predictions. Figure~\ref{fig5} shows the overall model architecture. 
    
    For each positive query triplet, multiple reliable logical reasoning paths are extracted correspondingly. After being formatted into sentences, triplets and the corresponding paths have similar semantics, which are measured by the cosine similarity:
    \begin{equation}
    \footnotesize
        \textit{cos}(s_1,s_2)=\frac{s_1 \cdot s_2}{||s_1||_2 \cdot ||s_2||_2},
    \end{equation}
    where $s_1$ and $s_2$ are the corresponding sentence embeddings.

    KRST extracts at most $|M|$ paths per triplet and converts paths and triplets into embeddings. The similarity score between each path and the corresponding triplet is computed and the path with the highest score is deemed as most reasonable for relation prediction. Therefore, we use the highest score among all paths as the score for each triplet:

    \begin{equation}
    \footnotesize
        \textit{score}(h,r,t)= \max_{p \in P} \{ \textit{cos}(s((h,r,t)),s(p)) \},
    \end{equation}
    where $P$ denotes the corresponding path set w.r.t triplet $(h,r,t)$, and $s(\cdot)$ denotes the embedding function (i.e., KRST) for triplets and paths.

    During training, negative triplets are processed with the label of -1. As aforementioned, commonly applied binary loss functions (e.g., cross-entropy and hinge) require the negative scores to be close to -1 (or 0). This is not appropriate in our scenario because the unmatched pair of triplet and path are not necessarily perpendicular to each other. To relax the penalty for loss, we use the cosine embedding loss function:
    \begin{equation}
    \label{eq_cos_loss}
    \footnotesize
        \mathscr{L}(s_1,s_2,y)=\left\{
        \begin{aligned}
        & 1-\textit{cos}(s_1,s_2), &  y=1,\\
        & \max(0,\textit{cos}(s_1,s_2)-\textit{margin}), & y=-1,
        \end{aligned}
        \right.
    \end{equation}
    where $\textit{margin} \in (-1,1)$ and $y \in \{1,-1\}$ denotes the label.
    Equation~\ref{eq_cos_loss} makes the positive score to be close to 1, maximizing similar semantics. On the other hand, the negative score is only constrained to be smaller than $\textit{margin}$. 
    
    \begin{table}[!t]
        \footnotesize
        \centering
        \caption{Hyper-parameter values of KRST}\label{table_config}
        \begin{tabular}{lcc}
        \toprule
        Description                                                                                                                & \multicolumn{2}{c}{Value}     \\ \midrule \midrule
        Random seed                                                                                                                & \multicolumn{2}{c}{42}        \\
        Samples for training and validation                                                                                        & \multicolumn{2}{c}{5}         \\
        Samples for testing                                                                                                        & \multicolumn{2}{c}{50}        \\
        \begin{tabular}[c]{@{}l@{}}Number of paths extracted for training,\\  validation and testing ($M$)\end{tabular}            & \multicolumn{2}{c}{3}         \\
        Path extraction search depth ($L$)                                                                                         & \multicolumn{2}{c}{5}         \\
        \begin{tabular}[c]{@{}l@{}}Support type \\ (1:coverage, 2:confidence).\end{tabular}                                        & 1            & 2              \\
        \begin{tabular}[c]{@{}l@{}}Support threshold \\ ($\textit{cover}_{R_{p^{'}}}$ or $\textit{conf}_{R_{p^{'}}}$)\end{tabular} & $10^{-5}$    & $5*10^{-3}$    \\
        Number of epochs                                                                                                           & \multicolumn{2}{c}{30}        \\
        Sentence transformer learning rate                                                                                        & \multicolumn{2}{c}{$10^{-5}$} \\ \bottomrule
        \end{tabular}
    \end{table}
\section{Experiments}\label{section_experiments}

    We evaluate the performance of KRST in three different settings: transductive, inductive, and few-shot. Then, we demonstrate multi-aspect comprehensive explanations by clustering the embeddings of paths generated by KRST.
    We implement KRST\footnote{\scriptsize  https://github.com/ZhixiangSu/KRST} with a SOTA sentence transformer (all-mpnet-base-v2\footnote{\scriptsize https://huggingface.co/sentence-transformers/all-mpnet-base-v2}) on a Tesla V100 GPU with 16GB RAM. We do not fine tune the hyper-parameter values of KRST in different experiments. Instead, we apply the same set of hyper-parameter values in KRST (see Table~\ref{table_config}) for all datasets in all transductive, inductive, and few-shot settings.

    Following the evaluation tasks conducted in \cite{GRAIL} and \cite{BERTRL}, in this paper, we measure the rank and hit rate of one positive triplet among 49 negative triplets. We only randomly generate negative triplets and use them for training and validation. For a fair comparison, we use the negative triplets provided by \cite{BERTRL} for testing.
    
    \subsection{Datasets}
    To evaluate the transductive and inductive performance of KRST, we use all three datasets adopted in \cite{BERTRL}, which were introduced by \cite{GRAIL}\footnote{\scriptsize https://github.com/kkteru/grail}. These datasets are commonly adopted by various inductive approaches and they are the respective subsets of WN18RR, FB15k-237, and NELL-995. In the inductive setting, training entities have no overlap with testing entities.

        \begin{table}[tbh!]
    \footnotesize
        
        \centering
        \caption{Statistics of datasets}\label{table_dataset}
        \begin{tabular}{lllll}
        \toprule
        Dataset   & $G$               & $|R_G|$       & $|E_G|$      & \#Triplets \\ \midrule \midrule
        WN18RR    & train             & 9           & 2746       & 6670       \\
                  & train-2000        & 9           & 1970       & 2002       \\
                  & train-1000        & 9           & 1362       & 1001       \\
                  & test-transductive & 7           & 962        & 638       \\
                  & test-inductive    & 8           & 922        & 1991       \\ \midrule
        FB15k-237 & train             & 180         & 1594       & 5223       \\
                  & train-2000        & 180         & 1280       & 2008       \\
                  & train-1000        & 180         & 923        & 1027       \\
                  & test-transductive & 102         & 550        & 492        \\
                  & test-inductive    & 142         & 1093       & 2404       \\
                  \midrule
        NELL-995  & train             & 88          & 2564       & 10063      \\
                  & train-2000        & 88          & 1346       & 2011       \\
                  & train-1000        & 88          & 893        & 1020       \\
                  & test-transductive & 60          & 1936       & 968        \\
                  & test-inductive    & 79          & 2086       & 6621       \\ 
                  \bottomrule
        \end{tabular}
    \end{table}

    \renewcommand\arraystretch{0.77}
    \begin{table*}[tbh!]
    \caption{Transductive and inductive relation prediction results}\label{table_transductive_and_inductive}
    \centering
    \scriptsize
    \begin{tabular}{llcccccc}\toprule
          &                   & \multicolumn{3}{c}{Transductive}                 & \multicolumn{3}{c}{Inductive}                    \\ \cmidrule(lr){3-5} \cmidrule(lr){6-8}
          &                   & WN18RR         & FB15k-237      & NELL-995       & WN18RR         & FB15k-237      & NELL-995       \\ \midrule \midrule
    MRR   & RuleN             & 0.669          & 0.674          & 0.736          & 0.780          & 0.462          & 0.710          \\
          & GRAIL             & 0.676          & 0.597          & 0.727          & 0.799          & 0.469          & 0.675          \\
          & MINERVA           & 0.656          & 0.572          & 0.592          & -              & -              & -              \\
          & TuckER            & 0.646          & 0.682          & 0.800          & -              & -              & -              \\
          & KG-BERT           & -              & -              & -              & 0.547          & 0.500          & 0.419          \\
          & BERTRL            & 0.683          & 0.695          & 0.781          & 0.792          & 0.605          & \textbf{0.808} \\ \cmidrule(lr){2-8}
          & KRST (No filter)  & 0.881          & 0.671          & 0.730          & 0.883          & 0.713          & 0.753          \\
          & KRST (Coverage)   & 0.897          & 0.709          & \textbf{0.803} & \textbf{0.902} & 0.704          & 0.696          \\
          & KRST (Confidence) & \textbf{0.899} & \textbf{0.720} & 0.800          & 0.890          & \textbf{0.716} & 0.769          \\
          \midrule \midrule
    Hit@1 & RuleN             & 0.646          & 0.603          & 0.636          & 0.745          & 0.415          & 0.638          \\
          & GRAIL             & 0.644          & 0.494          & 0.615          & 0.769          & 0.390          & 0.554          \\
          & MINERVA           & 0.632          & 0.534          & 0.553          & -              & -              & -              \\
          & TuckER            & 0.600          & 0.615          & \textbf{0.729} & -              & -              & -              \\
          & KG-BERT           & -              & -              & -              & 0.436          & 0.341          & 0.244          \\
          & BERTRL            & 0.655          & 0.620          & 0.686          & 0.755          & 0.541          & \textbf{0.715} \\ \cmidrule(lr){2-8}
          & KRST (No filter)  & 0.807          & 0.576          & 0.618          & 0.803          & \textbf{0.602} & 0.633          \\
          & KRST (Coverage)   & 0.831          & 0.624          & 0.692          & \textbf{0.835} & 0.573          & 0.554          \\
          & KRST (Confidence) & \textbf{0.835} & \textbf{0.639} & 0.694          & 0.809          & 0.600          & 0.649   \\
          \bottomrule
    \end{tabular}
    \end{table*}
    
    \renewcommand\arraystretch{0.77}
    \begin{table*}[tbh!]
    \scriptsize
    \caption{Few-shot transductive and inductive relation prediction results}\label{table_few_shot}
    \centering
    \begin{tabular}{llcccccccccccc}\toprule
          &                   & \multicolumn{6}{c}{Transductive}                                                                    & \multicolumn{6}{c}{Inductive}                                                                       \\ \cmidrule(lr){3-8} \cmidrule(lr){9-14}
          &                   & \multicolumn{2}{c}{WN18RR}      & \multicolumn{2}{c}{FB15k-237}   & \multicolumn{2}{c}{NELL-995}    & \multicolumn{2}{c}{WN18RR}      & \multicolumn{2}{c}{FB15k-237}   & \multicolumn{2}{c}{NELL-995}    \\  \cmidrule(lr){3-4} \cmidrule(lr){5-6} \cmidrule(lr){7-8} \cmidrule(lr){9-10} \cmidrule(lr){11-12} \cmidrule(lr){13-14}
          &                   & 1000           & 2000           & 1000           & 2000           & 1000           & 2000           & 1000           & 2000           & 1000           & 2000           & 1000           & 2000           \\ \midrule \midrule
    MRR   & RuleN             & 0.567          & 0.625          & 0.434          & 0.577          & 0.453          & 0.609          & 0.681          & 0.773          & 0.236          & 0.383          & 0.334          & 0.495          \\
          & GRAIL             & 0.588          & 0.673          & 0.375          & 0.453          & 0.292          & 0.436          & 0.652          & 0.799          & 0.380          & 0.432          & 0.458          & 0.462          \\
          & MINERVA           & 0.125          & 0.268          & 0.198          & 0.364          & 0.182          & 0.322          & -              & -              & -              & -              & -              & -              \\
          & TuckER            & 0.258          & 0.448          & 0.457          & 0.601          & 0.436          & 0.577          & -              & -              & -              & -              & -              & -              \\
          & KG-BERT           & -              & -              & -              & -              & -              & -              & 0.471          & 0.525          & 0.431          & 0.460          & 0.406          & 0.406          \\
          & BERTRL            & 0.662          & 0.673          & 0.618          & 0.667          & 0.648          & 0.693          & 0.765          & 0.777          & 0.526          & 0.565          & 0.736          & \textbf{0.744} \\ \cmidrule(lr){2-14}
          & KRST (Confidence) & \textbf{0.871} & \textbf{0.882} & \textbf{0.696} & \textbf{0.701} & \textbf{0.743} & \textbf{0.781} & \textbf{0.886} & \textbf{0.878} & \textbf{0.679} & \textbf{0.680} & \textbf{0.745} & 0.738          \\ \midrule \midrule
    Hit@1 & RuleN             & 0.548          & 0.605          & 0.374          & 0.508          & 0.365          & 0.501          & 0.649          & 0.737          & 0.207          & 0.344          & 0.282          & 0.418          \\
          & GRAIL             & 0.489          & 0.633          & 0.267          & 0.352          & 0.198          & 0.342          & 0.516          & 0.769          & 0.273          & 0.351          & 0.295          & 0.298          \\
          & MINERVA           & 0.106          & 0.248          & 0.170          & 0.324          & 0.152          & 0.284          & -              & -              & -              & -              & -              & -              \\
          & TuckER            & 0.230          & 0.415          & 0.407          & 0.529          & 0.392          & 0.520          & -              & -              & -              & -              & -              & -              \\
          & KG-BERT           & -              & -              & -              & -              & -              & -              & 0.364          & 0.404          & 0.288          & 0.317          & 0.236          & 0.236          \\
          & BERTRL            & 0.621          & 0.637          & 0.517          & 0.583          & 0.526          & 0.582          & 0.713          & 0.731          & 0.441          & 0.493          & 0.622          & 0.628          \\ \cmidrule(lr){2-14}
          & KRST (Confidence) & \textbf{0.790} & \textbf{0.810} & \textbf{0.611} & \textbf{0.602} & \textbf{0.628} & \textbf{0.678} & \textbf{0.811} & \textbf{0.793} & \textbf{0.537} & \textbf{0.524} & \textbf{0.637} & \textbf{0.629} \\ \bottomrule
    \end{tabular}
    \end{table*}
    
    
    For few-shot evaluation, we use the corresponding few-shot transductive and few-shot inductive datasets given in \cite{BERTRL}. The statistics of all datasets used in this paper are presented in Table~\ref{table_dataset}.

    \subsection{Transductive and Inductive Relation Prediction}
    
    In transductive cases, we extract paths in the training graph for all training, validation, and testing triplets. However, in inductive cases, paths for testing triplets are not available from the training graph, because entities used for testing do not appear in the training graph (see Definition~\ref{def_inductive_relation_prediction}). Instead, we use the inductive graphs given in \cite{GRAIL} for path extraction. The detailed configurations for path extraction are presented in Table~\ref{table_config}.
    
    We benchmark the performance of KRST against SOTA inductive methods, SOTA embedding-based methods, and SOTA reinforcement learning methods.
    Table~\ref{table_transductive_and_inductive} shows the results of transductive and inductive relation prediction. Compared with SOTA methods, KRST methods achieve the best performance under $\textit{MRR}$ (5 of 6) and  $\textit{Hit@1}$ (4 of 6) metrics. Specifically, KRST methods achieve significant improvement in the transductive case of WN18RR (+0.216 for $\textit{MRR}$ and +18.0\% for $\textit{Hit@1}$), inductive case of WN18RR (+0.103 for $\textit{MRR}$ and +6.6\% for $\textit{Hit@1}$) and inductive case of FB15k-237 (+0.111 for $\textit{MRR}$ and +5.9\% for $\textit{Hit@1}$). 
    
    Among KRST methods, the majority of the best results are achieved by KRST with relation path confidence (8 of 12). Only in 1 of 12 cases, KRST with no filter performs the best. So we empirically show that filtering (especially relation path confidence) elevates model performance. As for the relatively inferior performance of relation path coverage, this is because a long path usually leads to an exponentially increasing number of reachable entities, making the relation path coverage filtering prefer shorter paths. However, short paths are not necessarily reliable. Therefore, KRST's performance is limited by unreliable paths unfiltered by relation path coverage.

    \subsection{Few-shot Relation Prediction}
    
    In the few-shot settings, wherein only subsets of the entire datasets are given for training, we conduct similar path extractions as done for the entire datasets. Specifically, for transductive cases, training, validation, and testing paths are all extracted from the entire training graph. For inductive cases, paths for inductive training are extracted from the entire training graph, while paths for inductive validation and inductive testing are both extracted from the inductive graph.
    
    Because KRST with relation path confidence achieves the best performance on the entire datasets, we apply it for all few-shot settings. As shown in Table~\ref{table_few_shot}, KRST with relation path confidence outperforms SOTA methods in 11 of 12 cases. The average transductive and inductive improvement for $\textit{MRR}$ and $\textit{Hit@1}$ is 0.119 and 0.082 (+3.1\% and +5.0\%), respectively. In addition, the performance gap between few-shot-1000 and few-shot-2000 samples is small (smaller than 0.015 for $\textit{MRR}$ in 5 of 6 cases). These results illustrate the strong generalization capability of KRST, which requires a lesser amount of training samples to achieve on-par performance.
    

    \begin{figure*}[!t]
        \centering
        
        \subfigure[Clusters for triplet  (Atlantic City, Film Genre, Drama).]{
            \includegraphics[scale=0.3]{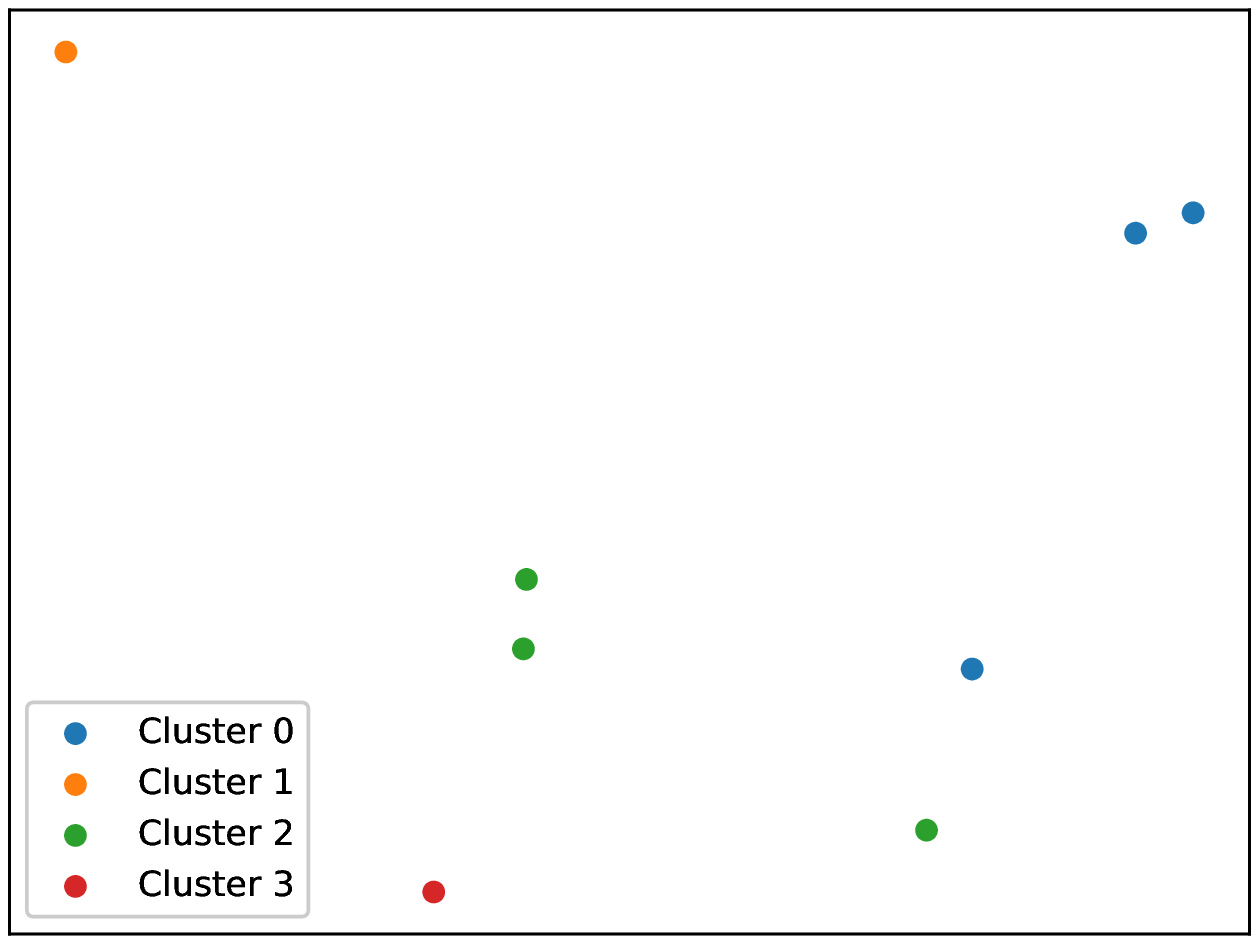}
            \label{fig_cluster2.1}
            }
        \quad
        \subfigure[Clusters for triplet (Kelsey Grammer, Person Language, English Language).]{
            \includegraphics[scale=0.3]{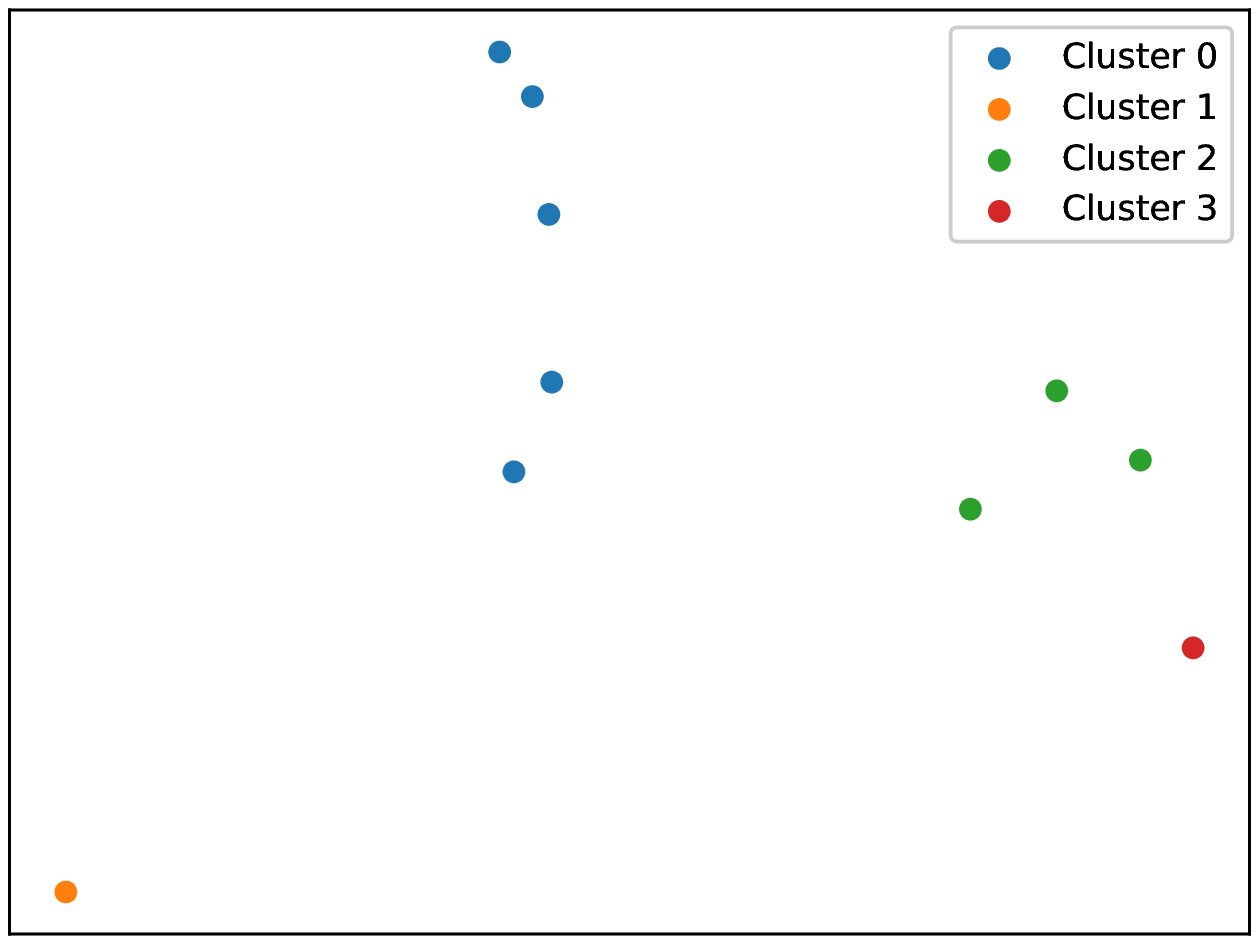}
            \label{fig_cluster2.3}
            }
        \quad
        \subfigure[Clusters for triplet (California, Location Contain, Carlsbad).]{
            \includegraphics[scale=0.3]{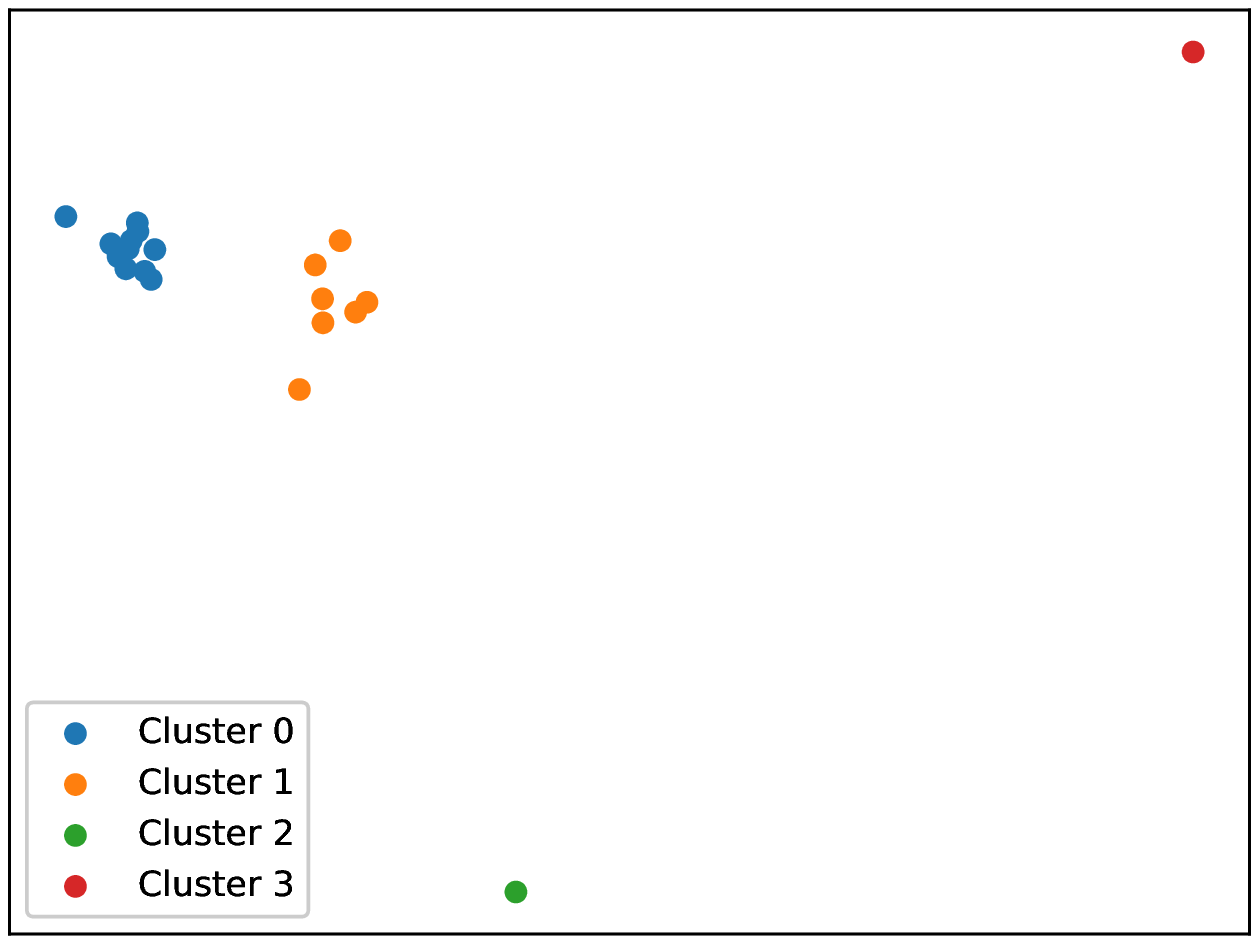}
            \label{fig_cluster2.2}
            }
        \caption{Clustering result for a multi-aspect explanation.}\label{pyfig0}
    \end{figure*}
    \begin{figure*}[tbh!]
            \centering
            \includegraphics[scale=0.75]{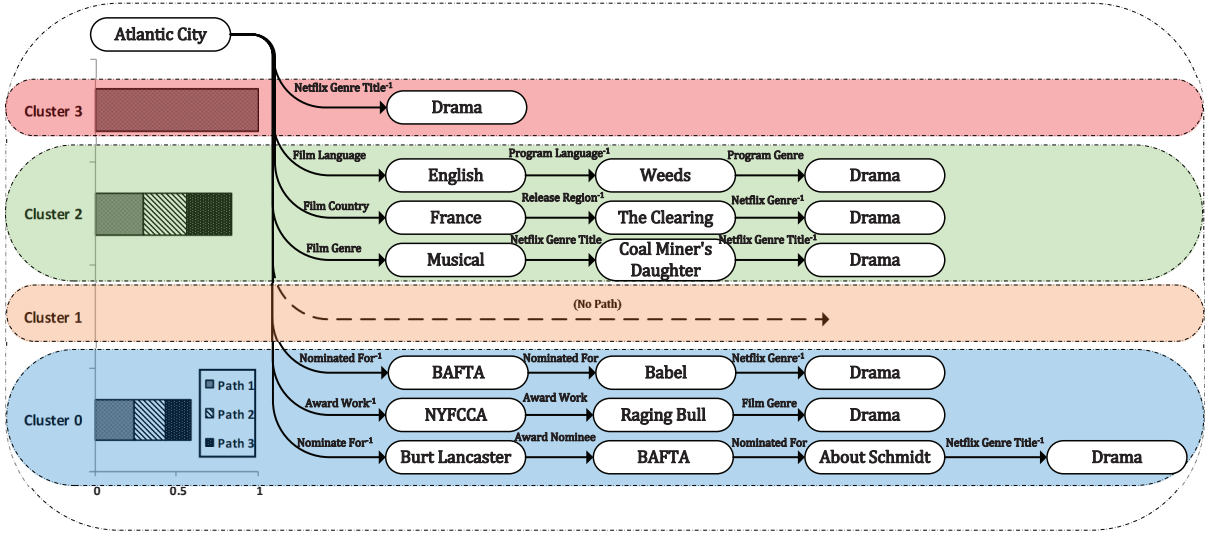}
            \caption{Paths clustered by similarity scores for a multi-aspect explanation of triplet (Atlantic City, Film Genre, Drama).}\label{fig6}
        \end{figure*}
    
    \subsection{Multi-Aspect Explanation}

    As afore-introduced, KRST is able to provide a multi-aspect explanation of the relation prediction results. This is because KRST generates an embedding for each path, allowing us to quantifiably analyze the relations among paths. Although paths are supposed to be similar to the given query triplet, they are not necessarily similar to each other. By grouping them into different clusters, we could provide a multi-aspect comprehensive explanation.

    Figure~\ref{pyfig0} visualizes the clustering results on the reduced dimensions after applying Linear Discriminant Analysis (LDA) for three different triplets. Clusters are generated using the K-Means algorithm. As shown in Figures~\ref{fig_cluster2.1} and~\ref{fig6}, KRST successfully provides a multi-aspect explanation for triplet $(\textit{AtlanticCity},\textit{FilmGenre},\textit{Drama})$ based on the similarity score (pre-processed after min-max scaling). The paths grouped into each cluster are shown in Figure~\ref{fig6} and presented using the same color in Figure~\ref{fig_cluster2.1}. As shown, the only path in Cluster~3 illustrates the explanation provided by the external knowledge from Netflix that $\textit{AtlanticCity}$ is a $\textit{Drama}$ available on Netflix. This explanation is straightforward and convincing, and is considered to be the most reliable (score of 1.0). For paths in Clusters 2, explanations are provided using similar films with the same attributes (e.g., language and country), which is similar to the human analogical reasoning. Cluster~0 explains the target triplet using the knowledge of awards won or nominated regarding $\textit{AtlanticCity}$ in three paths, which is an important and distinctive piece of side information.
    KRST considers Clusters~2 and 0 as relatively less convincing and assigns relatively lower scores (0.837 and 0.593 on average, respectively).
    We also input an empty path for comparison, which individually constitutes Cluster~1. KRST correctly gives it the lowest similarity score of 0. In summary, based on Figure~\ref{fig6}, we can easily perceive different explanations in the corresponding aspects of Netflix platform, similar films, and awards. More examples of the clustered paths for multi-aspect explanations are presented in Supp-Tables~\ref{table_appendix_multi_aspect_1} and~\ref{table_appendix_multi_aspect_2}.

    \section{Conclusion}
    In this paper, we propose a novel architecture named KRST which outperforms SOTA models in most transductive and inductive relation prediction tasks (15 of 18). In addition, we perform clustering on KRST generated embeddings and provide a comprehensive multi-aspect explanation.
    Nonetheless, KRST is a model based on BERT, which requires relatively large memory usage and computational resources.
    Going forward, we plan to solve this computational intensive issue by proposing a more parameter-efficient model.
    
    \section{Acknowledgement}
    This research is supported, in part, by the Joint SDU-NTU Centre for Artificial Intelligence Research (C-FAIR), Shandong University, China and by the Joint NTU-WeBank Research Centre on Fintech (Award No: NWJ-2020-010), Nanyang Technological University, Singapore. This work is also supported, in part, by the National Key R\&D Program of China No.2021YFF0900800; NSFC No.91846205; Shandong Provincial Key Research and Development Program (Major Scientific and Technological Innovation Project) (NO.2021CXGC010108).

\bibliography{ref}

\begin{appendix}
\onecolumn
\section{Appendix}
\subsection{More on Multi-Aspect Explanation}
We provide more case studies for multi-aspect explanation, which are shown in  Supp-Tables~\ref{table_appendix_multi_aspect_2}, and \ref{table_appendix_multi_aspect_1}. Paths grouped in different clusters are split using horizontal lines. The visualizations of the clustered paths shown in Supp-Tables~\ref{table_appendix_multi_aspect_2} and \ref{table_appendix_multi_aspect_1} are presented in Figures 3(b) and 3(c) in the submitted paper, respectively.

\begin{table*}[tbh!]
\footnotesize
\centering
\caption{Clustered paths for triplet (Kelsey Grammer, person language, English Language) with confidence scores}\label{table_appendix_multi_aspect_2}
\begin{tabular}{llc}
\toprule
Triplet                                                                                                                                                 & Path                                                                                                                                                                                                                                                                                                                                                                                                                                   & Score      \\ \midrule
\begin{tabular}[c]{@{}l@{}}$\textit{Kelsey Grammer}$    \\ $\xrightarrow[]{\textit{person language}}$    \\ $\textit{English Language}$\end{tabular}    & \begin{tabular}[c]{@{}l@{}}$\textit{Kelsey Grammer}  \xrightarrow[]{\textit{type of union}}  \textit{Marriage}  \xrightarrow[]{\textit{spouse person}}  \textit{Diane Sawyer} $\\ $   \xrightarrow[]{\textit{study of field}}  \textit{\textit{English Language}} $\end{tabular}                                                                                                                                                        & 0.577 \\
& \begin{tabular}[c]{@{}l@{}}$\textit{Kelsey Grammer}  \xrightarrow[]{\textit{nominee award}}  \textit{Peter Casey}  \xrightarrow[]{\textit{person profession}}  \textit{Screenwriter} $\\ $   \xrightarrow[]{\textit{profession person}}  \textit{Diane Sawyer}  \xrightarrow[]{\textit{study of field}}  \textit{English Language} $\end{tabular}                                                                                       & 0.580 \\
& \begin{tabular}[c]{@{}l@{}}$\textit{Kelsey Grammer}  \xrightarrow[]{\textit{nomination award nominee}}  \textit{David Angell}  \xrightarrow[]{\textit{person profession}}    \textit{Screenwriter}$\\ $  \xrightarrow[]{\textit{profession person}}  \textit{Diane Sawyer}  \xrightarrow[]{\textit{study of field}}  \textit{English Language} $\end{tabular}                                                                           & 0.561 \\\cmidrule{2-3}
& \begin{tabular}[c]{@{}l@{}}$\textit{Kelsey Grammer}  \xrightarrow[]{\textit{person gender}}  \textit{Male}  \xrightarrow[]{\textit{gender person}}  \textit{Philosophy} $\\ $   \xrightarrow[]{\textit{study of field}}  \textit{English Language} $\end{tabular}                                                                                                                                                                       & 0.653 \\\cmidrule{2-3}
& \begin{tabular}[c]{@{}l@{}}$\textit{Kelsey Grammer}  \xrightarrow[]{\textit{person religion}}  \textit{Catholicism}  \xrightarrow[]{\textit{religion person}}  \textit{Brittany Snow} $\\ $   \xrightarrow[]{\textit{dubbing performance language}}  \textit{English Language} $\end{tabular}                                                                                                                                                    & 1.000 \\
& \begin{tabular}[c]{@{}l@{}}$\textit{Kelsey Grammer}  \xrightarrow[]{\textit{nomination award}}  \textit{SAGA}  \xrightarrow[]{\textit{nominee award}}  \textit{Will Arnett}$\\ $  \xrightarrow[]{\textit{dubbing performance language}}    \textit{English Language} $\end{tabular}                                                                                                                                                                       & 0.484 \\
& \begin{tabular}[c]{@{}l@{}}$\textit{Kelsey Grammer}  \xrightarrow[]{\textit{nominee award}}  \textit{Jay Kogen}  \xrightarrow[]{\textit{tv program}}  \textit{The Simpsons}$\\ $  \xrightarrow[]{\textit{tv program language}}    \textit{English Language} $\end{tabular}                                                                                                                                                              & 0.774 \\
& \begin{tabular}[c]{@{}l@{}}$\textit{Kelsey Grammer}  \xrightarrow[]{\textit{nomination award}}  \textit{Tony Award for Best Actor in a Musical}  \xrightarrow[]{\textit{nominee award}}    \textit{Tim Curry}$\\ $  \xrightarrow[]{person language}  \textit{English Language} $\end{tabular}                                                                                                                                           & 0.780 \\
& \begin{tabular}[c]{@{}l@{}}$\textit{Kelsey Grammer}  \xrightarrow[]{\textit{student education graduates}}  \textit{Juilliard School}   \xrightarrow[]{\textit{education student}}  \textit{Tim Blake Nelson}$\\ $  \xrightarrow[]{\textit{actor film performance film}}    \textit{The Thin Red Line}  \xrightarrow[]{\textit{film language}}  \textit{English Language} $\end{tabular}                                                 & 0.572
\\ \cmidrule{2-3}
& (No Path) & 0.000\\ \bottomrule
\end{tabular}
\end{table*}
\begin{table*}[tbh!]
\footnotesize
\centering
\caption{Clustered paths for triplet $(\textit{California}, \textit{LocationContain}, \textit{Carlsbad})$ with confidence scores}\label{table_appendix_multi_aspect_1}
\begin{threeparttable}
\begin{tabular}{llc}
\toprule
Triplet                                                                                                                                                 & Path                                                                                                                                                                                                                                                                                                                                                                                                                                   & Score       \\ \midrule
\begin{tabular}[c]{@{}l@{}}$\textit{California}$\\ $\xrightarrow[]{\textit{location contain}}$    \\ $\textit{Carlsbad}$\end{tabular} & $\textit{California} \xrightarrow[]{\textit{state location bibs}} \textit{Carlsbad}$                                                                                                                                                                                                                                                                                                                                                    & 0.390 \\ \cmidrule{2-3}
& $\textit{California} \xrightarrow[]{\textit{location contain}} \textit{San Diego County}  \xrightarrow[]{\textit{location contain}} \textit{Carlsbad}$                                                                                                                                                                                                                                                                                  & 1.000 \\ \cmidrule{2-3}
& $\textit{California}  \xrightarrow[]{\textit{ceremony of location}}  \textit{Marriage}  \xrightarrow[]{\textit{location of ceremony}}  \textit{Carlsbad} $                                                                                                                                                                                                                                                                              & 0.801 \\
& \begin{tabular}[c]{@{}l@{}}$\textit{California}  \xrightarrow[]{\textit{location contain}}  \textit{Mountain View}  \xrightarrow[]{\textit{ceremony of location}}  \textit{Marriage}$\\ $  \xrightarrow[]{\textit{location of ceremony}}  \textit{Carlsbad} $\end{tabular}                                                                                                                                                              & 0.639 \\
& \begin{tabular}[c]{@{}l@{}}$\textit{California}  \xrightarrow[]{\textit{location contain}}  \textit{Santa Clara}  \xrightarrow[]{\textit{ceremony of location}}  \textit{Marriage} $\\ $ \xrightarrow[]{\textit{location of ceremony}}  \textit{Carlsbad} $\end{tabular}                                                                                                                                                                & 0.649 \\
& \begin{tabular}[c]{@{}l@{}}$\textit{California}  \xrightarrow[]{\textit{location contain}}  \textit{La Jolla}  \xrightarrow[]{\textit{ceremony of location}}  \textit{Marriage}  $\\ $\xrightarrow[]{\textit{location of ceremony}}  \textit{Carlsbad} $\end{tabular}                                                                                                                                                                   & 0.575 \\
& \begin{tabular}[c]{@{}l@{}}$\textit{California}  \xrightarrow[]{\textit{location contain}}  \textit{San Pedro}  \xrightarrow[]{\textit{ceremony of location}}  \textit{Marriage}  $\\ $\xrightarrow[]{\textit{location of ceremony}}    \textit{Carlsbad} $\end{tabular}                                                                                                                                                                & 0.621 \\
& \begin{tabular}[c]{@{}l@{}}$\textit{California}  \xrightarrow[]{\textit{lived places person}}  \textit{Sharon Osbourne}  \xrightarrow[]{\textit{type of union}}  \textit{Marriage}  $\\ $\xrightarrow[]{\textit{location of ceremony}}    \textit{Carlsbad} $\end{tabular}                                                                                                                                                              & 0.617 \\
& \begin{tabular}[c]{@{}l@{}}$\textit{California}  \xrightarrow[]{\textit{state location bibs}}  \textit{Sunnyvale}  \xrightarrow[]{\textit{location adjoin}}  \textit{Mountain View}  $\\ $  \xrightarrow[]{\textit{ceremony of location}}  \textit{Marriage}  \xrightarrow[]{\textit{location of ceremony}}  \textit{Carlsbad} $\end{tabular}                                                                                           & 0.310 \\\cmidrule{2-3}
& \begin{tabular}[c]{@{}l@{}}$\textit{California}  \xrightarrow[]{\textit{office government}}  \textit{Member of Congress}  \xrightarrow[]{\textit{government politician}}  \textit{James Madison}  $\\ $  \xrightarrow[]{\textit{type of union}}  \textit{Marriage}  \xrightarrow[]{\textit{location of ceremony}}  \textit{Carlsbad} $\end{tabular}                                                                                     & 0.327 \\
& \begin{tabular}[c]{@{}l@{}}$\textit{California}  \xrightarrow[]{\textit{statistical region religions}}  \textit{Christianity}  \xrightarrow[]{\textit{religion person}}  \textit{Will Smith} $\\ $   \xrightarrow[]{\textit{type of union}}  \textit{Marriage}  \xrightarrow[]{\textit{location of ceremony}}  \textit{Carlsbad} $\end{tabular}                                                                                         & 0.549 \\
& \begin{tabular}[c]{@{}l@{}}$\textit{California}  \xrightarrow[]{\textit{lived places person}}  \textit{Vivica A. Fox}  \xrightarrow[]{\textit{award winner}}  \textit{Will Smith} $\\ $   \xrightarrow[]{\textit{type of union}}  \textit{Marriage}  \xrightarrow[]{\textit{location of ceremony}}  \textit{Carlsbad} $\end{tabular}                                                                                                    & 0.449 \\
& \begin{tabular}[c]{@{}l@{}}$\textit{California}  \xrightarrow[]{\textit{mailing headquarters organization}}  \textit{Sony Pictures Entertainment}    \xrightarrow[]{\textit{nominee award}}  \textit{Jay-Z}$\\ $  \xrightarrow[]{\textit{type of union}}  \textit{Marriage}  \xrightarrow[]{\textit{location of ceremony}}  \textit{Carlsbad} $\end{tabular}                                                                            & 0.344 \\
& \begin{tabular}[c]{@{}l@{}}$\textit{California}  \xrightarrow[]{\textit{represented district}}  \textit{104th United States Congress}  \xrightarrow[]{\textit{government politician}}    \textit{John Kerry}$\\ $  \xrightarrow[]{\textit{type of union}}  \textit{Marriage}  \xrightarrow[]{\textit{location of ceremony}}  \textit{Carlsbad} $\end{tabular}                                                                                    & 0.356 \\
& \begin{tabular}[c]{@{}l@{}}$\textit{California}  \xrightarrow[]{\textit{mailing headquarters organization}}  \textit{California State University}  \xrightarrow[]{\textit{graduates education student}}    \textit{Kevin Costner}$\\ $  \xrightarrow[]{\textit{type of union}}  \textit{Marriage}  \xrightarrow[]{\textit{location of ceremony}}  \textit{Carlsbad} $\end{tabular}                                                                                    & 0.361 \\
& \begin{tabular}[c]{@{}l@{}}$\textit{California}  \xrightarrow[]{\textit{mailing headquarters organization}}  \textit{Pixar}  \xrightarrow[]{\textit{nominated for}}    \textit{Monsters}$\\ $  \xrightarrow[]{\textit{film release region}}  \textit{France}  \xrightarrow[]{\textit{ceremony of location}} \textit{Marriage} \xrightarrow[]{\textit{location of ceremony}} \textit{Carlsbad} $\end{tabular}                                                                                    & 0.356 \\
& \begin{tabular}[c]{@{}l@{}}$\textit{California}  \xrightarrow[]{\textit{burial of place}}  \textit{Clark Gable}  \xrightarrow[]{\textit{person gender}}  \textit{Male} $\\ $ \xrightarrow[]{\textit{gender person}}  \textit{Jeff Moss}    \xrightarrow[]{\textit{type of union}}  \textit{Marriage}  $\\ $\xrightarrow[]{\textit{location of ceremony}}  \textit{Carlsbad} $\end{tabular}                                              & 0.562 \\
& \begin{tabular}[c]{@{}l@{}}$\textit{California}  \xrightarrow[]{\textit{burial of place}}  \textit{Samuel Goldwyn}  \xrightarrow[]{\textit{person gender}}  \textit{Male} $\\ $ \xrightarrow[]{\textit{gender person}}  \textit{Jeff Moss}    \xrightarrow[]{\textit{type of union}}  \textit{Marriage}  $\\ $\xrightarrow[]{\textit{location of ceremony}}  \textit{Carlsbad} $\end{tabular}                                           & 0.552 \\
& \begin{tabular}[c]{@{}l@{}}$\textit{California}  \xrightarrow[]{\textit{lived places person}}  \textit{Frank Gehry}  \xrightarrow[]{\textit{person gender}}  \textit{Male} $\\ $ \xrightarrow[]{\textit{gender person}}  \textit{Jeff Moss}    \xrightarrow[]{\textit{type of union}}  \textit{Marriage}  $\\ $\xrightarrow[]{\textit{location of ceremony}}  \textit{Carlsbad} $\end{tabular}                                          & 0.491 \\
& \begin{tabular}[c]{@{}l@{}}$\textit{California}  \xrightarrow[]{\textit{mailing headquarters organization}}  \textit{Genentech}  \xrightarrow[]{\textit{service language}}  \textit{\textit{English Language}}$\\ $    \xrightarrow[]{\textit{language person}}  \textit{Aidan Gillen}  \xrightarrow[]{\textit{type of union}}  \textit{Marriage}  $\\ $\xrightarrow[]{\textit{location of ceremony}}  \textit{Carlsbad} $\end{tabular} & 0.296 \\ \bottomrule
\end{tabular}
\begin{tablenotes}
\item[] We adopt an automatic approach that if 20 paths are extracted, we do not add in the empty path, denoted as ``(no path)'', for clustering (as reflected in the source code), which is the case shown in this table.
\end{tablenotes}
\end{threeparttable}
\end{table*}

\end{appendix}

\end{document}